\documentclass{article} %
\usepackage{iclr2020_conference,times}

\usepackage{amsmath,amsfonts,bm}

\def\eqref#1{equation~\ref{#1}}

\def\1{\bm{1}}

\DeclareMathAlphabet{\mathsfit}{\encodingdefault}{\sfdefault}{m}{sl}
\SetMathAlphabet{\mathsfit}{bold}{\encodingdefault}{\sfdefault}{bx}{n}

\DeclareMathOperator*{\argmax}{arg\,max}
\DeclareMathOperator*{\argmin}{arg\,min}

\DeclareMathOperator{\sign}{sign}

\usepackage{hyperref}
\usepackage{url}
\usepackage{amsmath}
\usepackage{amsthm}
\usepackage{amsfonts}
\usepackage{mathtools}
\usepackage{verbatim}
\usepackage{cleveref}
\usepackage{booktabs} %
\usepackage[inline]{enumitem}
\usepackage{wrapfig}
\usepackage{multirow}
\usepackage{algorithm}
\usepackage{algorithmic}
\usepackage{abraces}

\theoremstyle{definition}
\newtheorem{defn}{Definition}[section]

\DeclareMathOperator*{\maximize}{maximize}
\DeclareMathOperator*{\minimize}{minimize}

\DeclareMathOperator*{\subjectto}{subject\;to}

\DeclareMathOperator*{\proj}{proj}
\DeclareMathOperator*{\step}{step}

    \title{Improved Wasserstein Attacks and Defenses}

\author{Edward J. Hu, Greg Yang, Adith Swaminathan \& Hadi Salman \\
Microsoft Research\\
Redmond, WA 98052, USA \\
\texttt{edward@edwardjhu.com}\\
\texttt{\{gregyang,adswamin,Hadi.Salman\}@microsoft.com}
}

\iclrfinalcopy %
\begin{document}

\maketitle

\begin{abstract}
Robustness of deep computer vision models against image perturbations bounded by a $\ell_p$ ball have been well-studied recently. Perturbations in the real-world, however, rarely exhibit the pixel independence that $\ell_p$ threat models assume.
A recently proposed Wasserstein distance-bounded threat model is a promising alternative that limits the perturbation to pixel mass movements.
We point out and rectify flaws in the previous definition of the Wasserstein threat model and explore stronger attacks and defenses under our better-defined framework.
Lastly, we discuss the inability of current Wasserstein-robust models in defending against perturbations seen in the real world.
We will release our code and trained models upon publication.
\end{abstract}

\section{Introduction}

\enlargethispage{\baselineskip}

Deep learning approaches to computer vision tasks, such as image classification, are not robust.
For example, a data point that is classified correctly can be modified in a nearly imperceptible way to cause the classifier to misclassify it \citep{szegedy2013intriguing, goodfellow2014explaining}.
Projected Gradient Descent (PGD) is a well-studied method to find such small perturbations within a $\ell_p$ ball of a small radius \citep{madry2017towards}.
While both general and effective, the $\ell_p$
threat model perturbs each pixel independently, a property not seen in realistic perturbations, such as distortion, blurring, and spatial shifts.

Previously, Wasserstein distance has been proposed as a more perceptually-aligned metric for images~\citep{peleg_unified_1989}.
Recently, \citet{wong_wasserstein_2019} proposed a Wasserstein distance-based threat model as an alternative to the $\ell_p$ threat model, and derived a computationally feasible approach to project onto the Wasserstein ball during PGD.
\autoref{fig:illu} (Left) shows a perturbation found when attacking a $\ell_p$ robust model with our Wasserstein threat model.
The perturbation looks different from the one found by a $\ell_p$ threat model.

The Wasserstein threat model of \citet{wong_wasserstein_2019} provided a great foundation, but only considered \emph{normalized} images; their attack algorithm, when applied to real images, could produce a perturbed image that is \textit{outside} the allowed Wasserstein ball.
In this work, we define the Wasserstein threat model such that it applies to all images, and we provide a \textit{safe} algorithm to find adversarial perturbations within a specified Wasserstein radius.
Our algorithm uses a \textit{constrained} Sinkhorn iteration to project images onto the intersection of a Wasserstein ball and a $\ell_\infty$ ball; the computational overhead from our new constraint is offset by our run-time optimizations and justified by our stronger attacks.
Further, we provide a significantly stronger attack than \citet{wong_wasserstein_2019}'s by exploring different PGD steps, which we incorporate in an adversarial training framework to obtain more robust models.

The main contributions of our work include:
\begin{itemize}
    \item A definition of the Wasserstein adversarial threat model for all images of the same dimensionality, fixing a glitch in the previous formulation (\autoref{sec:wasserstein_threat});
    \item A constrained Sinkhorn iteration projection algorithm that produces adversarial examples under our new, better-formulated threat model (\autoref{sec:constrained_sinkhorn});
    \item A significantly stronger Wasserstein-bounded attack by taking a different PGD step; our new attack breaks adversarially trained Wasserstein-robust models in prior works (\autoref{sec:attacks});
    \item An adversarially trained model that achieves state-of-the-art robustness against the new attacks, while still being robust against the attacks in prior works (\autoref{sec:defense}).
\end{itemize}

\begin{table*}[t]
    \centering
    \small
    \begin{tabular}{llrrrrrrrr} \toprule
         MNIST & $1$-Wasserstein $\epsilon \times n_{pixel}$ & 5 & 10 & 20 & 50 & 100 & 200 & 500 & 1000 \\ \midrule
         & \citet{wong_wasserstein_2019} (\%) & 100 & 100 & 100 & 100 & 94 & 91 & 83 & 71\\
         & Our Attack (\%)  & 100 & 90 & 85 & 68 & 39 & 8 & 1 & 1\\ \midrule
         CIFAR-10 & $1$-Wasserstein $\epsilon \times n_{pixel}$ & 5 & 10 & 20 & 50 & 100 & 200 & 500 & 1000 \\ \midrule
         & \citet{wong_wasserstein_2019} (\%) & 82 & 82 & 82 & 82 & 82 & 81 & 80 & 77\\
         & Our Attack (\%) & 76 & 68 & 58 & 35 & 20 & 9 & 1 & 0\\ \bottomrule
    \end{tabular}
    \caption{Empirical top-1 accuracy of the adversarially trained Wasserstein-robust classifier from~\citet{wong_wasserstein_2019} under their and our attack at various Wasserstein radii ($\epsilon$) multiplied by the per image pixel count $n_{pixel}$. Results are for MNIST and CIFAR-10. Lower accuracy at a given $\epsilon$ suggests a stronger attack.}
    \label{tab:attack}
\end{table*}

\section{Adversarial Robustness for Images}

Deep neural networks (DNNs) can be fooled into making wrong predictions by deliberately changing the input in ways imperceptible to humans.
In this section we summarize common approaches for such attacks and defenses against them.

\paragraph{Adversarial Attacks}
One well-studied example is the \textit{additive $\ell_p$-bounded attack} \citep{madry2017towards, goodfellow2014explaining, carlini2017towards,wong2018provable}, where a perturbation designed to increase the loss of the correct label, usually found using first-order information, is added to the input.
PGD performs this addition and projection jointly and iteratively~\citep{madry2017towards}, which gives an efficient empirical algorithm for finding adversarial examples.
The $\ell_p$ norm of total perturbation is bounded to a small $\epsilon$ so that the change is imperceptible.
Another example is the \textit{pixel-wise functional attack} \citep{laidlaw2019functional}, where a function is applied to each pixel individually, with constraints on the function itself and/or on the output pixel, enabling attacks in the color space, for instance.
\textit{Image-wide attacks} can also be used to fool DNNs. These are defined by a function that applies to the image as a whole, such as translations and rotations, with $\ell_p$ constraints on the function parameters~\citep{mohapatra_towards_2019}.
In this work, we focus on a different class of attacks, specifically adversarial examples with a bounded Wasserstein distance~\citep{wong_wasserstein_2019}. %

\paragraph{Adversarial Defenses}

In order to defend a model against a given adversary, one can employ empirical or certified defenses. Empirical defenses are basically heuristics that are used to train robust models. One of the most effective empirical defenses is \textit{adversarial training}~\citep{goodfellow2014explaining, madry2017towards, wong_wasserstein_2019}, in which a given predictive model is trained on adversarial examples generated by a given threat model. This defense, although empirical, has proven to be one of the strongest defenses in practice. Certified defenses on the other hand are those that provide guarantees under a specific threat model~\citep{wong2018provable, cohen2019certified, salman_provably_2019, levine_wasserstein_2019, salman2019convex, weng2018towards,raghunathan2018certified}, but these often yield weaker empirical performance.
In this work, we focus on empirical defenses, specifically adversarial training.

\section{Wasserstein Distance-based Threat Model}
\label{sec:wasserstein_threat}

The $p$-Wasserstein distance between two probability distributions measures the ``minimal effort'' needed to rearrange the probability mass in one distribution so it matches the other one.
More formally, given two distributions $A$ and $B$ over a metric space $X$ with metric $d$, the $p$-Wasserstein distance is defined as $W_p(A,B)=\left[\inf_\Gamma \mathbb{E}_{(x,y)\sim \Gamma}  d(x,y)^p \right]^\frac{1}{p}$ where $\Gamma$ is a distribution over the product space $X \times X$ such that its marginals are $A$ and $B$.
Given two images represented as $3$-dimensional tensors, $x, x' \in [0, 1]^{m,n,c}$ where $m,n$ are the dimensions of the image and $c$ is the number of channels (e.g. for RGB $c=3$); we measure the $p$-Wasserstein distance between them by normalizing both tensors into probability distributions.
Intuitively, the $p$-Wasserstein distance measures the cost of transporting pixel mass to turn one image into the other\footnote{We only allow pixel mass movement within channels to limit our problem to finding the 2D Wasserstein distance, following~\citet{wong_wasserstein_2019}.}%
, with the transport costing pixel distance (measured in $d$) to the $p^{th}$ power per unit mass. %
We define the \emph{$p$-Wasserstein distance} of images $x$ and $x'$ of the same dimensionality with non-zero $\ell_1$-norm for all channels as:
\begin{equation}
    W_p(x, x') = \sum_{i \in \{R,G,B\}}{W_p\left(\frac{x_i}{||x_i||_1}, \frac{x'_i}{||x'_i||_1}\right)}
\end{equation}
Note that $x$ and $x'$ can have different $\ell_1$ norms.
According to~\citet{wong_wasserstein_2019}, an adversarial example $x'$ of radius $\epsilon$ under the $p$-Wasserstein threat model is one that causes the neural network to give a different prediction than that of $x$ and also satisfies: $W_p(x, x') \leq \epsilon.$

\begin{figure}[t]
    \centering
    \includegraphics[width=0.34\textwidth]{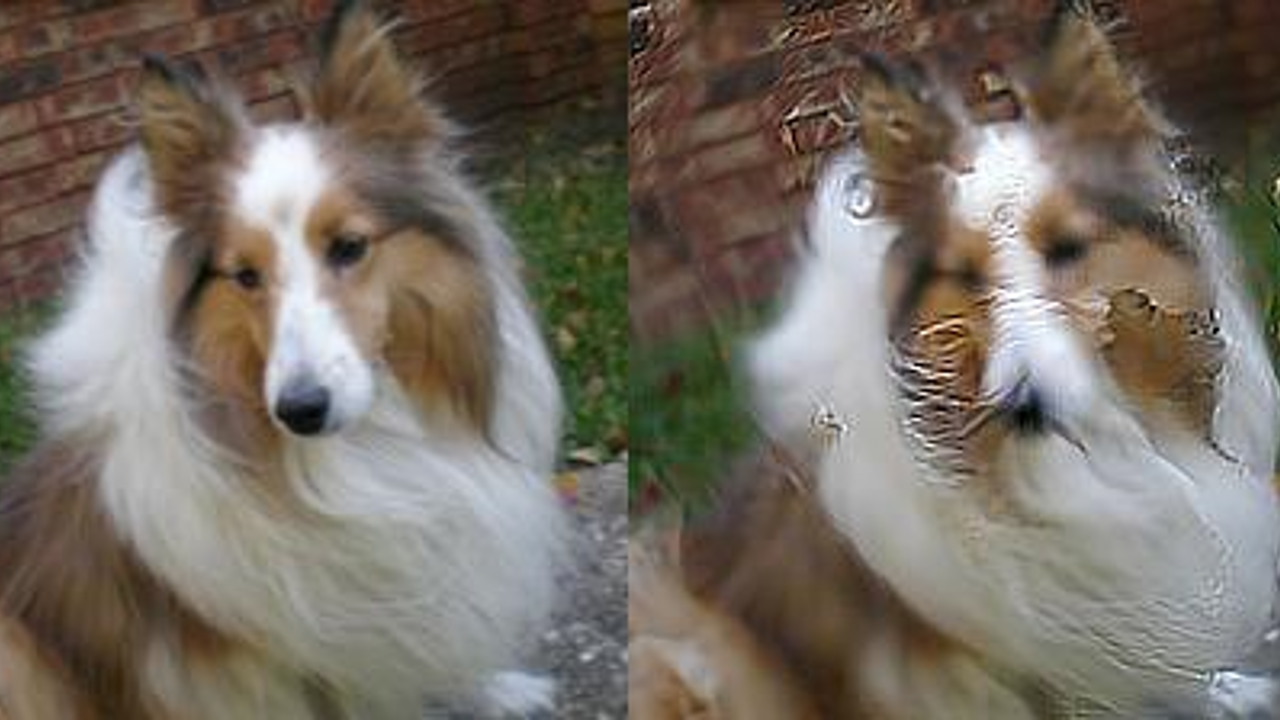}
    \includegraphics[width=0.32\textwidth]{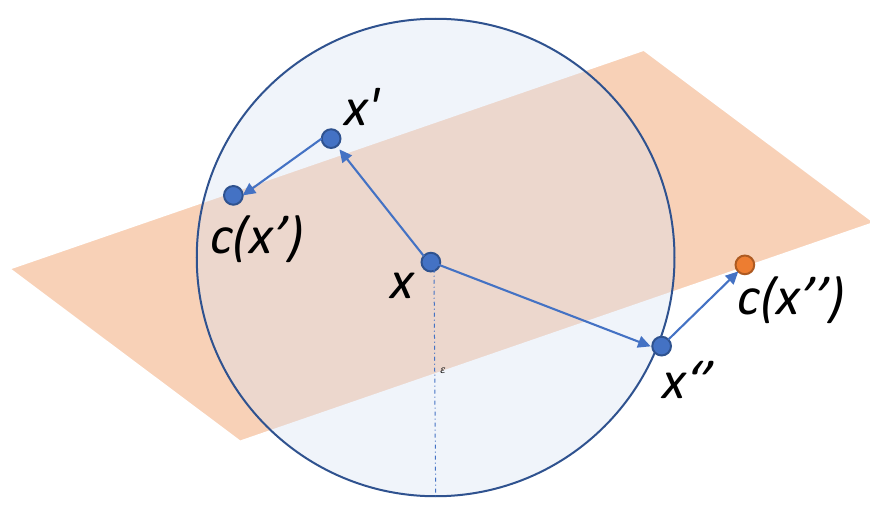}
    \includegraphics[width=0.32\textwidth]{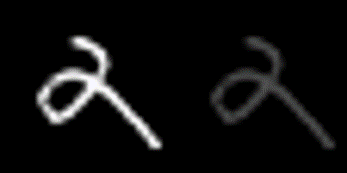}
    \caption{\textbf{Left}: A Wasserstein perturbation found by our attack that causes the $\ell_2$ robust model from~\citet{engstrom_adversarial_2019} to misclassify a image from ImageNet~\citep{imagenet}.
    \textbf{Middle}: Clamping the perturbed image ($x', x''$) can increase the Wasserstein distance to the original image ($x$), especially it is if near the boundary; the blue circle represents a  Wasserstein ball of radius $\epsilon$; the orange parallelogram represents an $\ell_\infty$ ball; $c(\cdot)$ is the clamping function.
    \textbf{Right}: We break the adversarially trained Wasserstein-robust model from~\citet{wong_wasserstein_2019} by dimming the image brightness, which does not change the normalized image. The example here has its brightness reduced by 3x.}
    \label{fig:illu}
\end{figure}

\paragraph{An Obvious Flaw} This threat model only considers the probability distribution represented by the image but not the total pixel mass (i.e. the $\ell_1$-norm of the image, reflected as brightness).
We devise a trivial attack to break the Wasserstein-robust model from~\citet{wong_wasserstein_2019} by simply dimming the brightness of images from the MNIST test set. By dividing the image tensor by $30$ ($x \mapsto x/30$), we cause the model to misclassify $86$\% of the test set, even though all the dimmed data points have a Wasserstein distance of 0 to the original undimmed data points, as illustrated in \autoref{fig:illu} (Right).

Intuitively, a threat model that allows pixel mass movements should preserve the total pixel mass.
Therefore, we further require our Wasserstein threat model to preserve the $\ell_1$ norm, or the total pixel mass, after the perturbation.
This is a commonly acknowledged constraint to make the Wasserstein distance a true metric~\citep{rubner_earth_2000} and was implicit in the attack-finding PGD algorithm of ~\citet{wong_wasserstein_2019}, where the original image's $\ell_1$ norm is used to re-scale their Sinkhorn projected probability distributions.

\begin{defn}
For a radius $\epsilon$ and an image $x$, we define our \emph{constrained Wasserstein ball} to be
\begin{align}
    \mathcal{B}(x, \epsilon)=\{x' : W_p(x, x') \leq \epsilon, \|x\|_1 = \|x'\|_1, \overbrace{0 \leq x' \leq 1}^{\text{pixel range $[0, 1]$}} \}
    \label{eq:W-ball}
\end{align}
An adversarial example of $f$ in an $\epsilon$ neighborhood of $x$ is any $x' \in \mathcal{B}(x, \epsilon)$ such that $f(x') \neq f(x)$.
\end{defn}

\section{Constrained Sinkhorn Iteration}
\label{sec:constrained_sinkhorn}

\paragraph{PGD Iteration}
During untargeted\footnote{We maximize the loss of the correct label during an untargeted attack, and minimize the loss of a particular incorrect label during a targeted attack.} PGD attacks, one updates the input iteratively as follows:

\begin{align}
    x^{(t+1)} = \proj_{\mathcal{B}(x, \epsilon)}(x^{(t)}+\alpha \step(\nabla \ell(f(x^{(t)}), y)))
    \label{eq:PGD}
\end{align}

where $\proj_{\mathcal{B}(x, \epsilon)}=\argmin_{x' \in \mathcal{B}(x, \epsilon)}{\|x-x'\|_2^2}$ where $\mathcal{B}(x, \epsilon)$ is determined by the threat model.
$step$ is a function that takes in the gradient of the model $f$ with respect to $x^{(t)}$ and a step size $\alpha$, and outputs a step which is added to $x^{(t)}$.

For example, $\ell_\infty$ PGD has
\begin{align}
    \step=\sign(\nabla \ell(f(x^{(t)}), y)), \; \mathcal{B}(x, \epsilon)=\{x' : \|x-x'\|_\infty \leq \epsilon , 0 \leq x' \leq 1 \}
    \label{eq:linf_PGD}
\end{align}
and $\ell_2$ PGD has
\begin{align}
   \step=\frac{\nabla \ell(f(x^{(t)}), y)}{\|\nabla \ell(f(x^{(t)}), y)\|_2}, \; \mathcal{B}(x, \epsilon)=\{x' : \|x-x'\|_2 \leq \epsilon , 0 \leq x' \leq 1 \}
   \label{eq:l2_PGD}
\end{align}

In this section, we detail our changes to the function $\proj_{\mathcal{B}(x, \epsilon)}$ where $\mathcal{B}(x, \epsilon)$ describes our threat model as in \autoref{eq:W-ball}.
We investigate the effect of the $step$ function in \autoref{sec:attacks}.

To summarize our change to the $\proj_{\mathcal{B}(x, \epsilon)}$ function, we
\begin{enumerate}
    \item add a $\ell_\infty$ constraint to eliminate the need for clamping;
    \item improve run-time by re-using the dual variables across PGD steps;
    \item modify termination conditions to allow an explicit trade-off between safety and efficiency.
\end{enumerate}
Critically, this algorithm allows us to safely explore stronger empirical attacks that find perturbations close to the allowed Wasserstein ball boundary around the original image.

\paragraph{The Original Algorithm in \cite{wong_wasserstein_2019}}
In the sequel, assume that all images are represented as vectors in $\mathbb{R}^n_+$ and bounded in $[0, 1]$. Projecting a perturbed image onto a Wasserstein ball around the original image requires solving the following optimal transport problem:
\begin{equation}
\minimize_{z\in\mathbb R^n_+, \Pi\in \mathbb{R}^{n\times n}_+} \;\; \frac{1}{2}\|w-z\|_2^2\quad
\mathrm{s.t.} \;\;  \Pi1 = x,\;\; \Pi^T1 = z,\;\;
 \langle \Pi,C\rangle \leq \epsilon
\label{eq:projection}
\end{equation}
$w$ is the image after taking a gradient step; $x$ is the original image; $\Pi$ is the transport plan; $C$ is the cost matrix; and $z$ is the projected image.
\citet{wong_wasserstein_2019} made this projection efficient by approximating with an entropy-regularized optimization problem, which is solved using Lagrange multipliers and coordinate-descent on the dual variables.

\paragraph{Clamping: the flaw of the algorithm in \cite{wong_wasserstein_2019}}
As shown in \autoref{eq:W-ball}, \autoref{eq:linf_PGD}, and \autoref{eq:l2_PGD}, we need to project the perturbed image into the valid pixel range.
$\ell_p$-based PGD attacks achieve this by clamping every pixel to $[0, 1]$ after projecting onto the $\ell_p$ ball.
This ad-hoc clamping is also used in many works studying PGD attacks using the Wasserstein threat model~\citep{wong_wasserstein_2019, levine_wasserstein_2019}.
However, clamping does not guarantee that the clamped image is still within the allowed Wasserstein radius, as illustrated in \autoref{fig:illu} (Middle).
Under the attack of~\citet{wong_wasserstein_2019}, the perturbed image often uses less than 50\% of the Wasserstein budget, and thus rarely goes outside of the Wasserstein ball despite this unsafe clamping.
However, as we derive stronger attacks that project points onto the boundary of the Wasserstein ball, ad-hoc clamping causes the resultant images to be over-budget by 50\% on average, and in some cases by over 200\% when using the original Sinkhorn iteration projection.
This shows that the need to clamp an un-normalized image introduces a critical flaw in~\citet{wong_wasserstein_2019}'s attack-finding algorithm.
We bypass this need for clamping by deriving a constrained Sinkhorn iteration to project directly onto the intersection of the Wasserstein ball and an $\ell_\infty$ ball, making this ad-hoc clamping unnecessary.

\paragraph{Our Algorithm}
As discussed in~\autoref{sec:wasserstein_threat}, a perturbed distribution should not only be within the Wasserstein ball of the original image after normalization but also be element-wise within $[0, 1]$ after un-normalization by multiplying the original $\ell_1$ norm $||x||_1$.
This can be expressed as an additional $\ell_\infty$ constraint on $z$: $0 \leq z \leq r, r = \frac{1}{||w||_1}$, which can be simplified to $z_j \leq r$ for $j =1, .., n$ since $z$ is already constrained to be non-negative.

The new entropy-regularized optimized problem with the new constraint is as follows:
\begin{equation}
\minimize_{z\in\mathbb R^n_+, \Pi\in \mathbb{R}^{n\times n}_+} \;\; \frac{\lambda}{2}\|w-z\|_2^2 + \sum_{ij}\Pi_{ij}\log(\Pi_{ij})\quad
\mathrm{s.t.} \;\; \Pi1 = x, \Pi^T1 = z, \langle \Pi,C\rangle \leq \epsilon,
z_j \leq r
\label{eq:constrained_sinkhorn}
\end{equation}
for $j = 1, ..., n.$
We introduce dual variables $(\alpha, \beta, \psi, \phi)$ where $\psi, \phi_j \geq 0$, for $j=0, ..., n$. The dual of the problem is:
\begin{gather*}
    \maximize_{\alpha, \beta \in\mathbb R^n_+, \psi \in \mathbb{R}_+, \phi\in \mathbb{R}^n_+}
    g(\alpha, \beta, \psi, \phi),\quad
    \text{where}\\
    g(\alpha, \beta, \psi, \phi)
    =
    \begin{cases}
    & - \frac{1}{2\lambda}\|\beta+\phi\|_2^2 - \psi \epsilon + \alpha^Tx + \beta^Tw + \phi^Tw \\
    &\quad - r\sum_j{\phi_j} - \sum_{ij}\exp(\alpha_i)\exp( - \psi C_{ij} - 1)\exp( \beta_j).
    \end{cases}
\end{gather*}
We leave the details of the derivation to \autoref{app:lagrangian}.
By maximizing $g$ w.r.t individual dual variables, we obtain the solution to the dual variables:
\begin{align}
    \argmax_{\alpha_i}g(\alpha,\beta,\psi,\phi)
    &= \log \left(x_i\right) - \log \left(\sum_{j}\exp( - \psi C_{ij} - 1)\exp( \beta_j)\right)
        \label{eq:sinkhorn_iterate}\\
    \argmax_{\beta_j}g(\alpha,\beta,\psi,\phi)
    &= \lambda w_j - \phi_j - W\left(\lambda \exp(-\phi_j+\lambda w_j)\sum_{i}\exp(\alpha_i - \psi C_{ij} - 1)\right)
        \label{eq:argmax_beta}
\end{align}
Note that~\autoref{eq:sinkhorn_iterate} is identical to the one in~\citet{wong_wasserstein_2019};~\autoref{eq:argmax_beta} has an additional variable $\phi_j$.
Since $L$ is quadratic in $\phi$ with a negative coefficient, the maximization w.r.t. $\phi$ yields:
\begin{equation*}
\begin{gathered}
\quad \argmax_{\phi_j}g(\alpha,\beta,\psi,\phi) = \max(0, \lambda(w_j-r)-\beta_j)
\end{gathered}
\end{equation*}
We also perform Newton steps following~\citet{wong_wasserstein_2019} to find the solution to $\psi$ iteratively.
The primal solutions to our optimal transport problem can be recovered as:
\begin{equation}
\Pi_{ij} = \exp(\alpha_i)\exp(-\psi C_{ij}-1)\exp(\beta_j),\quad
z  = -\lambda^{-1} (\beta+\phi) + w.
\end{equation}

\subsection{Run-time Optimization}
\label{app:optimization}

We initialize the dual variables to the stopping condition in the previous PGD step, instead of the general starting condition.
Empirically, this allows the algorithm to converge faster, which is especially helpful since the additional constraint require more iterations to converge.
The full description of our algorithm (Algorithm~\ref{alg:projected_sinkhorn}) can be found in \autoref{sec:changes_to_sinkhorn}.

Below we benchmark the attack run-time on MNIST test (\autoref{tab:runtime}) set using the original Sinkhorn iteration~\citep{wong_wasserstein_2019}, our constrained version with and without this optimization, and pairing the optimized constrained Sinkhorn with our stronger attack, which is described in~\autoref{sec:attacks}.

\begin{table*}[ht]
    \centering
    \small
    \begin{tabular}{lcccc} \toprule
         Attack   & \citet{wong_wasserstein_2019} & +Constrained Proj. & +Optim. & +Our Attack \\ \midrule
         Run-time & 34 mins & 78 mins & 38 mins & 38 mins \\
         Avg. $\epsilon$& 0.443 & 0.460 & 0.460 & 0.135 \\ \bottomrule
    \end{tabular}
    \caption{Run-time and effectiveness for different attacks on a standard MNIST model using a single NVIDIA Tesla P100 GPU. Each column adds a modification to the previous one, from left-to-right: constrained Sinkhorn projection, our runtime optimization, and our stronger attack.  Average $\epsilon$ is the mean Wasserstein radius needed to break the entire MNIST test set (the lower the better). The overhead from the additional constraint is offset by our run-time optimization. Our attack introduces little overhead.}
    \label{tab:runtime}
\end{table*}

\subsection{Termination Conditions}
\label{sec:termination}
To ensure that the perturbed image is sufficiently compliant with our Wasserstein threat model, we incorporate both the Wasserstein and $\ell_1$ norm constraints into the termination condition for Sinkhorn iterations.
Specifically, we calculate the Wasserstein radius over-budget $W_{over}$ and $\Delta\ell_1$, the deviation between the sum of the output distribution and $1$, once the algorithm has converged or run for at least a set number of iterations.
\begin{align}
    W_{over} = - \epsilon + \sum_{ij}{C_{ij} \exp(\alpha_i)\exp(-\psi C_{ij}-1)\exp(\beta_j)} \\
    \Delta\ell_1 = abs( 1 - \sum_j z_j )
\end{align}
We terminate the algorithm only when both quantities are sufficiently small. For attacks, we set the threshold for $d_{W-over}$ to be $0.01 \times \epsilon$ and for $\Delta\ell_1$ to be 0.01.
Note that for adversarial training, where such strict compliance of the constraints might not be necessary, one can use more lenient thresholds to speed-up the projection at the expense of strict compliance with the threat model.

\section{Stronger Empirical Attacks}
\label{sec:attacks}

Now we take a look at the choice of the $\step$ function during PGD.
The empirical attack proposed in~\citet{wong_wasserstein_2019} uses the steepest descent w.r.t. the $\ell_\infty$ norm as their PGD step, with a step size tied to absolute pixel values.

\paragraph{Step Size}
The PGD step size for $\ell_p$ threat models is usually defined in the pixel space.
On the other hand, Wasserstein threat models manipulate the underlying distribution of images, thus, it is more natural to define step sizes in the normalized distribution space.
We find a trade-off between effectiveness and run-time - a larger step size can result in a stronger attack up to a point but takes longer for Sinkhorn iteration to converge.
Empirically, we pick a step size of $0.06$ for our experiments, which limits the change to a single pixel to 6\% of the total pixel mass and is much larger compared to~\citet{wong_wasserstein_2019}.
We quantified the effect of step size in~\autoref{app:larger_steps}, which shows that it does not affect the effectiveness of the attack once it is sufficiently large.

\begin{figure}
    \centering
    \includegraphics[width=0.47\textwidth]{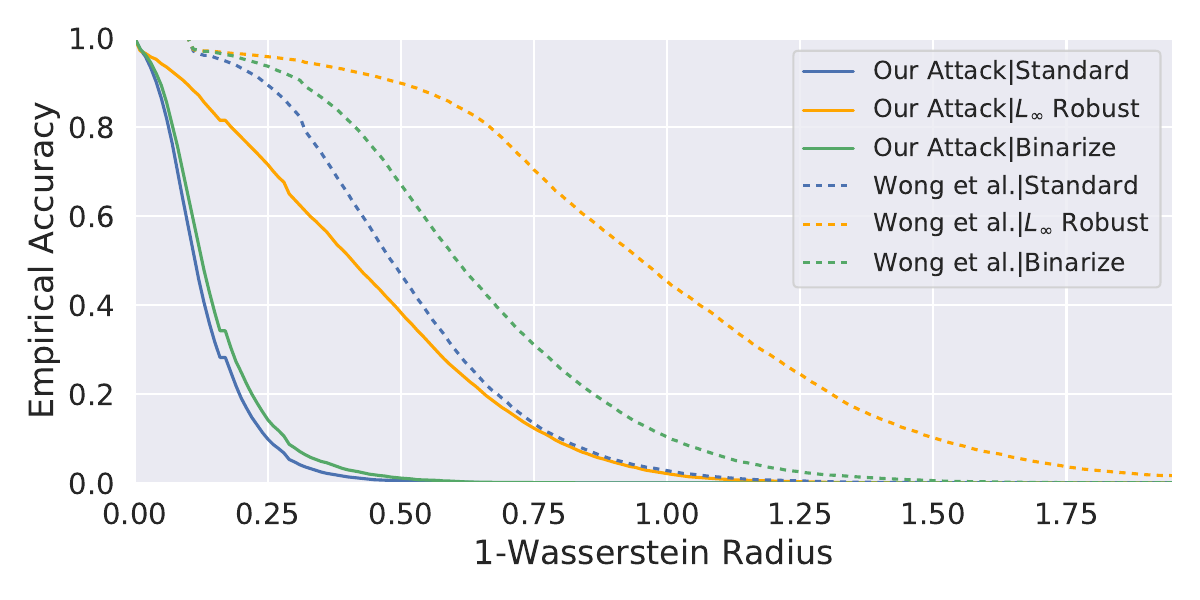}
    \includegraphics[width=0.47\textwidth]{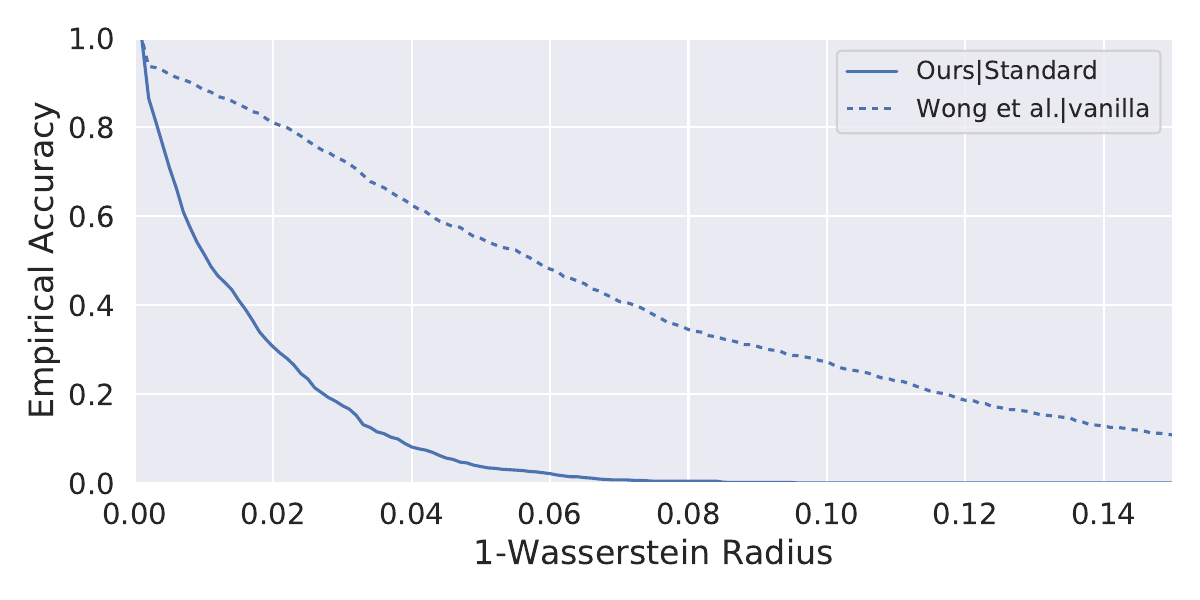}

    \caption{\textbf{Left}: Our new attack against the one in~\citet{wong_wasserstein_2019} on MNIST (lower is better). Given the same Wasserstein budget, we reduce the classifier accuracy significantly more.
    \textbf{Right}: the same result on CIFAR-10.}
    \label{fig:attacks}
\end{figure}

\paragraph{Gradient Step}
The additive perturbation before projection during PGD is a function of the gradient~\autoref{eq:PGD}, and~\cite{wong_wasserstein_2019} used the steepest descent w.r.t. the $\ell_\infty$ norm, which is effectively the sign of the gradient.
This is similar to the Fast Gradient Sign Method (FGSM)~\citep{goodfellow2014explaining} from the $\ell_p$ robustness literature, which is known to yield weaker attacks than the steepest descent w.r.t. the $\ell_2$ norm.
Motivated by this, we replace the $\step$ function with the steepest descent w.r.t. the $\ell_2$ norm, and match the $\ell_\infty$ norm of our gradient step and the original gradient step for a fair comparison.
The empirical accuracy of both our attack and the original one is shown in \autoref{fig:attacks}.

\paragraph{Ablation Study}

We ablate the effect of increasing the step size and using the steepest descent w.r.t. the $\ell_2$ norm (\autoref{tab:ablation}).
Our result shows that the improvement comes from both changes, and significant gain comes from using a different gradient step that considers not just the sign.

\begin{table*}[ht]
    \centering
    \small
    \begin{tabular}{llrrrrrrrr} \toprule
         MNIST & $1$-Wasserstein $\epsilon \times n_{pixel}$ & 5 & 10 & 20 & 50 & 100 & 200 & 500 & 1000 \\ \midrule
         & \citet{wong_wasserstein_2019} (\%) & 100 & 100 & 100 & 100 & 94 & 91 & 83 & 71\\
         & +large $\alpha$ (\%)  & 100 & 96 & 96 & 94 & 88 & 69 & 12 & 0\\
         & +large $\alpha$ + $\ell_2$ descent (\%)  & 100 & 90 & 85 & 68 & 39 & 8 & 1 & 1\\ \midrule
         CIFAR-10 & $1$-Wasserstein $\epsilon \times n_{pixel}$ & 5 & 10 & 20 & 50 & 100 & 200 & 500 & 1000 \\ \midrule
         & \citet{wong_wasserstein_2019} (\%) & 82 & 82 & 82 & 82 & 82 & 81 & 80 & 77\\
         & +large $\alpha$ (\%) & 81 & 80 & 78 & 70 & 55 & 23 & 1 & 0\\
         & +large $\alpha$ + $\ell_2$ descent (\%) & 76 & 68 & 58 & 35 & 20 & 9 & 1 & 0\\ \bottomrule
    \end{tabular}
    \caption{Ablation study on the effect of larger step size ($\alpha$) and using steepest descent w.r.t. the $\ell_2$ norm. The model under attack is the adversarially trained Wasserstein-robust model from~\citep{wong_wasserstein_2019}.}
    \label{tab:ablation}
\end{table*}

\section{Defending against a Wasserstein Adversary}
\label{sec:defense}

Following~\citet{wong_wasserstein_2019}, we use adversarial training to defend against our Wasserstein attack.
We train our MNIST model for 100 epochs and CIFAR-10 model for 200 epochs with a termination condition of $W_{over}=0.1$ and $\Delta\ell_1=0.1$ (introduced in \autoref{sec:termination}) for efficiency.
The model is trained on perturbed inputs with $\epsilon$ growing from $0.1$ to $10$ on an exponential schedule.

\begin{figure}[ht]
    \centering
    \includegraphics[width=0.47\textwidth]{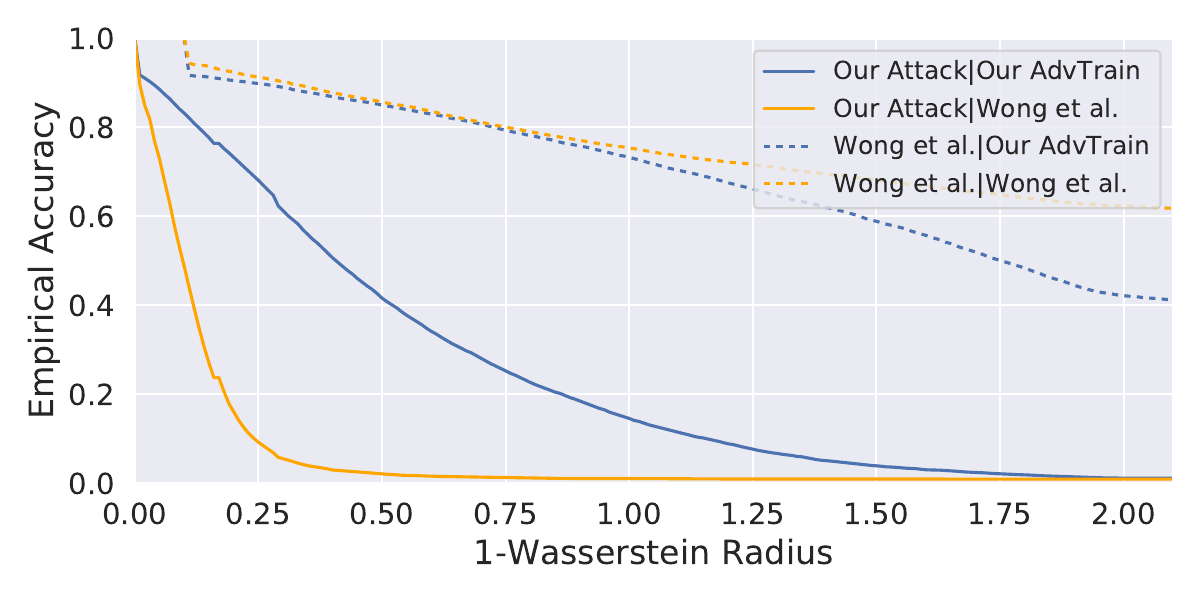}
    \includegraphics[width=0.47\textwidth]{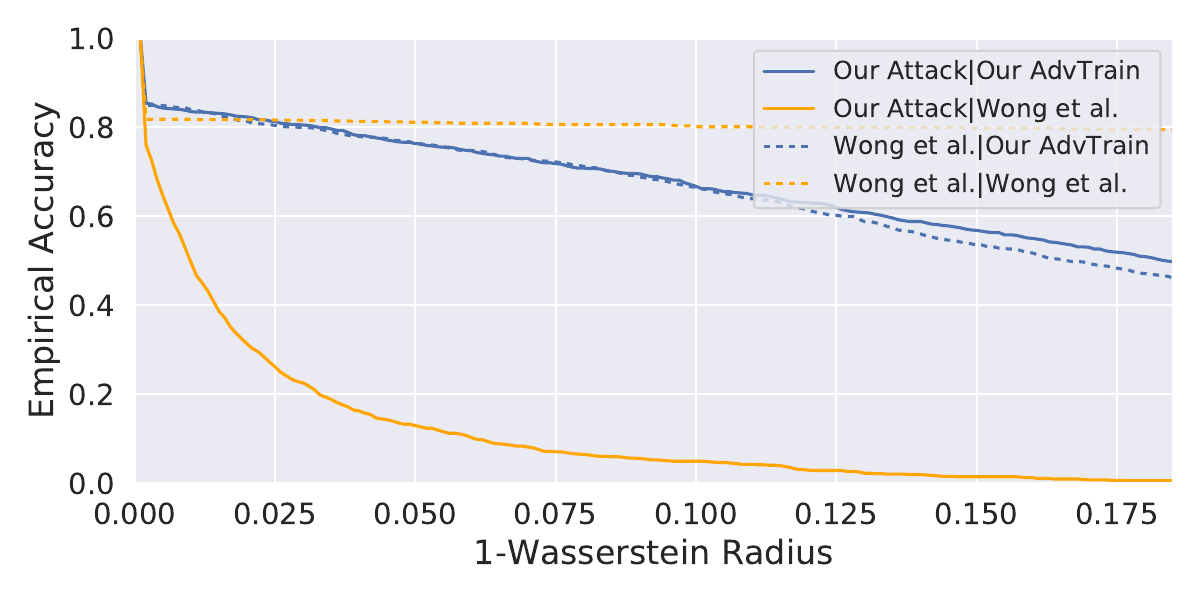}
    \caption{
    \textbf{Left}: Adversarial training under our attack and that of~\citet{wong_wasserstein_2019} on MNIST (higher is better). Our defended model is more robust to our attack, while also robust to the attack from~\citet{wong_wasserstein_2019}.
    \textbf{Right}: the same result on CIFAR-10} %
    \label{fig:defenses}
\end{figure}

We report our result in \autoref{fig:defenses}.
Our defended model is more robust against our stronger attack compared to the previously defended model from~\cite{wong_wasserstein_2019}, while still being generally robust against the attack in~\cite{wong_wasserstein_2019}.
Note that for large $\epsilon$, our defended model performs slightly worse than the previously defended model under the weaker attack.
We believe this could be that the adversarially trained model from~\citet{wong_wasserstein_2019} overfits the weaker attack, similar to the ``catastrophic overfitting" phenomenon~\citep{wong2020fast} associated with FGSM.
Another hypothesis is that we need to increase model capacity to defend against the stronger attack; however, We are not able to meaningfully reduce this gap by doubling the width of the layers in the model.

\section{Not Ready to Defend Against Common Natural Perturbations}
\label{sec:limitations}

The Wasserstein threat model is motivated by $\ell_p$ threat model's failure to bound common natural perturbations, e.g., translation, rotation, and blurring.
Such perturbations, when their magnitude is small, are barely perceptible to human, yet they incur a large change in the $\ell_p$ distance.
One reason behind this is the $\ell_p$ threat model perturbs each pixel independently, while the Wasserstein threat model does not.
We investigate empirically if a Wasserstein-robust model is more robust against translations, rotations, and blurring.
For some intuition regarding the $\ell_p$ and Wasserstein distance incurred by such perturbations, please see \autoref{app:common_distance}.

As it turns out, we can rarely defend against such common perturbations by adversarially training against a Wasserstein adversary, as shown in \autoref{tab:performance_under_perturbation}.
The only effective case is when defending against Gaussian blurs on CIFAR-10.
We reflect on this result from two perspectives.

\paragraph{Better Defenses}
The Wasserstein threat model is much less studied than that of $\ell_p$, and the naive defense we use, adversarial training, does not defend against a large enough radius.
Once we improve our ability to defend against a larger Wasserstein radius, we ought to gain robustness against the natural perturbations we explored in this section.
This motivates future work on defending against the Wasserstein threat model.

\paragraph{Constraints on the Threat Model}
Furthermore, we believe that perturbations exist on a spectrum of semantic specificity.
On one end, we have a narrow set of semantic perturbations such as translations and rotations.
We can defend against them effectively through data augmentation.
On the other end, we have highly expressive perturbation classes such as those bounded by a Wasserstein ball, which are harder to defend against, but do not require knowing which perturbations are likely to occur.
Orthogonal to finding stronger defenses, we can constrain the expressiveness of our threat model to make existing defenses more specific to the type of perturbations of interest.

\begin{table*}[t]
    \centering
    \small
    \begin{tabular}{crcccccccccc} \toprule
                                    &      & Clean & \multicolumn{3}{c}{Translation} & \multicolumn{3}{c}{Rotation} &  \multicolumn{3}{c}{Gaussian Blur} \\
                                    &      &  /  & 5\% & 10\% & 20\% & 5$^{\circ}$ & 10$^{\circ}$ & 20$^{\circ}$ & 3 & 5 & 7 \\ \midrule
         \multirow{2}{*}{MNIST}     & Standard (\%) & \textbf{98.9} & \textbf{97.8} & \textbf{85.8} & 48.3 & \textbf{98.5} & \textbf{97.7} & \textbf{91.9} & \textbf{98.6} & \textbf{93.1} & \textbf{62.4} \\
                                    & Our Robust (\%) & 92.8 & 91.2 & 82.3 & \textbf{49.3} & 91.0 & 88.7 & 79.1 & 88.0 & 75.1 & 58.4 \\ \midrule
         \multirow{2}{*}{CIFAR-10}  & Standard (\%) & \textbf{94.8} & \textbf{94.5} & \textbf{94.5} & \textbf{93.6} & \textbf{91.6} & \textbf{86.6} & 64.1 & 35.1 & 17.8 & 15.4 \\
                                    & Our Robust (\%) & 84.4 & 84.0 & 84.0 & 81.7 & 82.8 & 81.3 & \textbf{70.6} & \textbf{69.5} & \textbf{35.1} & \textbf{24.7} \\ \bottomrule
    \end{tabular}
    \caption{The top-1 classification accuracy of a standard and our Wasserstein-robust model under fixed common perturbations. Gaussian blur is parameterized by the width of the kernel in pixels. Higher is better.}
    \label{tab:performance_under_perturbation}
\end{table*}

For example, the common perturbations we study exhibit highly ``smooth" movements with a directional (for translations and rotations) or a radial (for Gaussian blurs) pattern.
This ``smoothness" constraint is also exhibited in perturbations such as lens distortions and refraction, where one would reasonably assume that the Wasserstein distance is more meaningful than $\ell_p$ due to the dependency on neighboring pixels.

We leave the rigorous formulation of such a ``smoothness" constraint to future works, and note a few ideas and challenges in~\autoref{app:additional_constraints}

\section{Conclusion and Future Work}

We introduced a better-defined threat model based on Wasserstein distance for all images of the same dimensionality.
A procedure naively ported over from the $\ell_p$ threat model could lead to adversarial examples that are not compliant with the Wasserstein threat model.
We fixed this by proposing a constrained version of the original Sinkhorn iteration algorithm, which projects onto the intersection of a Wasserstein ball and an $\ell_\infty$ ball directly.
This allowed us to safely explore attacks that come close to the allowed Wasserstein boundary.
By increasing the attack step size and using the steepest descent w.r.t. the $\ell_2$ norm of the gradient, we obtained a significantly stronger empirical Wasserstein attack that breaks previously defended models easily.
We defended against our attack with adversarial training, and showed that our defended model was generally robust against prior attacks.

Nonetheless, our defended model remained inadequate to defend against large, real-world perturbations, such as translations, rotations, and blurring.
This highlighted the need to design better defenses against the Wasserstein threat model, as current approaches did not yield a large enough defense radius.
On the other hand, we noted that these perturbations exhibited smoothness, and could potentially be better modeled if we introduce additional constraints to our threat model.
The constrained Sinkhorn formulation we develop may be of independent interest and offers a versatile tool to encode constraints on perturbations. We anticipate future work that incorporates other constraints in the threat model and efficiently optimizes them using the constrained Sinkhorn method.%

\section*{Broader Impact}

Many of the safety critical computer vision systems, including self-driving cars, are vulnerable to adversarial attacks.
Our work rectifies flaws in a recently proposed %
threat model %
and discovers much stronger attacks.
Our constrained projection framework can enable adding several other convex constraints that can capture common perturbations seen in the wild. %

However, our work has several limitations discussed in Sections~\ref{sec:defense} and \ref{sec:limitations}. We do not have certified defenses against these attacks, and the threat model does not capture other realistic ways in which adversaries may target computer vision systems. Consequently, we caution practitioners from a false sense of security when employing Wasserstein-adversary robust models.

\subsubsection*{Acknowledgments}
We thank Tony Duan, Eric Wong, Ashish Kapoor, Jerry Li, Ilya Razenshteyn, Jeremy Cohen and the anonymous reviewers for their helpful comments.

\newpage
\bibliography{main}
\bibliographystyle{iclr2020_conference}

\newpage

\appendix
\section{Appendix}

\subsection{Attacking with larger step sizes}
\label{app:larger_steps}
We show in \autoref{tab:trade-off} the effect of attacking our adversarially trained model with different step sizes.
The model under attack is the same as the one in \autoref{sec:defense}, which is trained against an adversary with a step size of $0.06$.
We notice the trade-off between run-time and attack effectiveness as we increase the attack step size.
Notably, increasing the step size beyond what the model was trained against does not break the model.
Empirically, Sinkhorn iterations as currently implemented run into numerical issues with step sizes larger than $0.08$.
We anticipate that future work can address this numerical stability issue.

When attacking within a small Wasserstein radius (e.g., $<0.06$), we also find it helpful to decrease the PGD step size accordingly.
Intuitively, if only less than 6\% of the total pixel mass is allowed to move by $1$ pixel, a step size of 0.06 (as we have chosen for our experiments), which allows a single pixel to move by 6\% of the total pixel mass, only makes convergence slow without making the attack more effective.
We empirically set the step size to $\min(\frac{\epsilon}{2}, \alpha)$, where $\epsilon$ is the Wasserstein radius and $\alpha$ is the step size.

\begin{table*}[ht]
    \centering
    \small
    \begin{tabular}{lccrrrrr} \toprule
         MNIST & Step Size & Run-time & Acc. (\%)@100 & 200 & 400 & 800 & 1600 \\ \midrule
         & 0.02 & 1.8 hrs & 84 & 72 & 49 & 21 & 2 \\
         & 0.04 & 2.1 hrs & 81 & 68 & 42 & 15 & 1 \\
         & 0.06 & 2.5 hrs & 81 & 67 & 41 & 14 & 1 \\
         & 0.08 & 3.0 hrs & 81 & 67 & 41 & 14 & 1 \\ \midrule
         CIFAR-10 & Step Size & Run-time & Acc. (\%)@100 & 200 & 400 & 800 & 1600 \\ \midrule
         & 0.02 & 3.7 hrs & 52 & 44 & 44 & 44 & 44 \\
         & 0.04 & 4.6 hrs & 52 & 43 & 43 & 43 & 43 \\
         & 0.06 & 4.9 hrs & 52 & 43 & 43 & 43 & 43 \\
         & 0.08 & 5.0 hrs & 51 & 42 & 42 & 42 & 42 \\ \bottomrule
    \end{tabular}
    \caption{The trade-off between run-time and attack effectiveness for MNIST and CIFAR-10. The accuracy is evaluated on our adversarially trained model (against an attack step size of 0.06) at various Wasserstein radii multiplied by the per image pixel count $n_{pixel}$. Run-time is recorded on a single NVIDIA Tesla P100 GPU.}
    \label{tab:trade-off}
\end{table*}

\subsection{Derivation of the Lagrangian}
\label{app:lagrangian}

\begin{proof}
For convenience, we multiply the objective by $\lambda$, expand the $L_\infty$ norm to individual pixels, and solve this problem instead:
 \begin{equation}
\begin{split}
\minimize_{z\in\mathbb R^n_+, \Pi\in \mathbb{R}^{n\times n}_+} &\;\; \frac{\lambda}{2}\|w-z\|_2^2 + \sum_{ij}\Pi_{ij}\log(\Pi_{ij})\\
\subjectto \;\; & \Pi1 = x\\
& \Pi^T1 = z\\
& \langle \Pi,C\rangle \leq \epsilon\\
& z_j \leq r, j = 1, ..., n.
\end{split}
\end{equation}
Here $r$ is the maximal pixel value.
Introducing dual variables $(\alpha, \beta, \psi, \vec{\phi})$ where $\psi, \phi_j \geq 0$, for j=0, ..., n, the Lagrangian is
\begin{equation}
\begin{split}
&L(z, \Pi, \alpha, \beta, \psi, \vec{\phi}) \\
= &\frac{\lambda}{2}\|w-z\|_2^2 + \sum_{ij}\Pi_{ij}\log(\Pi_{ij})+ \psi (\langle \Pi,C\rangle - \epsilon) \\
& +\sum_{j}{\phi_j (z_j - r)} + \alpha^T(x - \Pi1) + \beta^T(z - \Pi^T1).
\end{split}
\end{equation}
The KKT optimality conditions are now
\begin{equation}
\begin{split}
\frac{\partial L}{\partial \Pi_{ij}} &= \psi C_{ij} + (1 + \log(\Pi_{ij})) - \alpha_i - \beta_j = 0\\
\frac{\partial L}{\partial z_j} &= \lambda(z_j-w_j) + \beta_j + \phi_j= 0
\end{split}
\end{equation}
so at optimality, we must have
\begin{equation}
\begin{split}
\Pi_{ij} &= \exp(\alpha_i)\exp(-\psi C_{ij}-1)\exp(\beta_j)\\
z  &= -\frac{\beta+\phi}{\lambda} + w
\end{split}
\end{equation}
Plugging in the optimality conditions, we get
\begin{equation}
\begin{split}
& L(z^*,\Pi^*, \alpha, \beta, \psi, \phi) \\
= &\frac{\lambda}{2}\|w-z\|_2^2 + \sum_{ij}\Pi_{ij}\log(\Pi_{ij})+ \psi (\langle \Pi,C\rangle - \epsilon) \\
& +\sum_{j}{\phi_j (z_j - r)} + \alpha^T(x - \Pi1) + \beta^T(z - \Pi^T1) \\
=&\frac{-\|\beta\|_2^2-\|\phi\|_2^2}{2\lambda} - \psi \epsilon + \alpha^Tx + \beta^Tw + \phi^Tw - r\sum_j{\phi_j} \\ &-\frac{\beta^T\phi}{\lambda} - \sum_{ij}\exp(\alpha_i)\exp( - \psi C_{ij} - 1)\exp( \beta_j)\\
=& g(\alpha, \beta, \psi, \phi)
\end{split}
\end{equation}
so the dual problem is to maximize $g$ over $\alpha, \beta, \psi, \phi \geq 0, for j = 0, ..., n$.
\end{proof}

\subsection{Changes to the Sinkhorn iteration algorithm}
\label{sec:changes_to_sinkhorn}

We further make a few changes to the Sinkhorn iterations proposed by~\citet{wong_wasserstein_2019} to make the algorithm more efficient. First, note that the argmax
\begin{align*}
    \phi_j^*, \beta_j^* := \argmax_{\phi_j, \beta_j} g(\alpha, \beta, \psi, \phi)
\end{align*}
can be obtained in one pass.
First suppose the argmax $\phi_j^*$ is non-negative.
Then by solving $\partial g/ \partial \beta_j = \partial g/ \partial \phi_j = 0$ we have the solutions
\begin{align*}
    \beta_j^* &= \log r - \log \sum_i  \exp(\alpha_i)\exp(-\psi C_{ij} - 1) \\
    \phi_j^* &= \max(\lambda(w_j - r) - \beta_j^*, 0)
\end{align*}
If $\lambda(w_j - r) - \beta_j$ above is negative, then $\phi_j^*$ has to be 0, and therefore
\begin{align*}
    &\beta_j^* = \lambda w_j - W\left(\lambda \exp(\lambda w_j)\sum_{i}\exp(\alpha_i)\exp( - \psi C_{ij} - 1)\right)
\end{align*}

Our final algorithm can be described in the following pseudo-code (Algorithm~\ref{alg:projected_sinkhorn}).
\begin{algorithm}[ht]
   \caption{Projected Sinkhorn iteration to project $x$ onto the $\epsilon$ Wasserstein ball
   around $y$. We use $\cdot$ to denote element-wise multiplication. The $\log$ and $\exp$ operators also apply element-wise. }
\label{alg:projected_sinkhorn}
\begin{algorithmic}
   \STATE {\bfseries input:} $x,w\in \mathbb R^n, C\in \mathbb C^{n\times n}, \lambda \in \mathbb R$
   \STATE Initialize $\alpha_i, \beta_i \coloneqq \log(1/n)$ for $i=1,\dots,n$ and $\psi,\phi\coloneqq 1$
   \STATE $u,v \coloneqq \exp(\alpha), \exp(\beta)$
   \WHILE{$\alpha,\beta,\psi,\phi$ not converged}
   \STATE \textit{// update $K$}
   \STATE $K_\psi \coloneqq \exp(-\psi C - 1)$
   \STATE
   \STATE \textit{// block coordinate descent iterates}
   \STATE $\alpha \coloneqq \log(x) - \log(K_\psi v)$
   \STATE $u \coloneqq \exp(\alpha)$
   \STATE $\beta_1 \coloneqq \lambda w - \phi - W\left(u^T K_\psi\cdot \lambda \exp(\lambda w - \phi) \right)$
   \STATE $\beta_2 \coloneqq \log r - \log \sum_i  \exp(\alpha_i)\exp(-\psi C_{ij} - 1)$
   \STATE $\phi \coloneqq \max(\lambda (w - r) - \beta_1, 0)$
   \STATE $\beta \coloneqq where(\phi < 0, \beta_2, \beta_1)$
   \STATE $v \coloneqq \exp(\beta)$
   \STATE
   \STATE \textit{// Newton step}
   \STATE $g \coloneqq -\epsilon + u^T(C\cdot K_\psi)v$
   \STATE $h \coloneqq - u^T(C \cdot C \cdot K_\psi)v$
   \STATE
   \STATE \textit{// ensure }$\psi \geq 0$
   \STATE $\psi \coloneqq \max(\psi - g/h, 0)$
   \ENDWHILE
   \STATE {\bfseries return:} $w - (\beta+\phi)/\lambda$
\end{algorithmic}
\end{algorithm}

\subsection{Common Real-world Perturbations under \texorpdfstring{$\ell_p$}{Lp} and Wasserstein distance}
\label{app:common_distance}
We first measure the distance change under $\ell_p$ and $1$-Wasserstein when we apply such common perturbations; the result is in \autoref{tab:distance}.
The raw distances are not comparable between the two metrics, since the $\ell_2$ distance is on the pixel space, while $1$-Wasserstein is on the underlying distribution space.
We note that the $1$-Wasserstein distance incurred by such perturbations scales roughly linearly with the magnitude, while the $\ell_2$ distance does not.

\begin{table*}[ht]
    \centering
    \small
    \begin{tabular}{ccccccccccc} \toprule
                                    &           & \multicolumn{3}{c}{Translation} & \multicolumn{3}{c}{Rotation} &  \multicolumn{3}{c}{Gaussian Blur} \\
                                    &           & 5\% & 10\% & 20\% & 5$^{\circ}$ & 10$^{\circ}$ & 20$^{\circ}$ & 3 & 5 & 7 \\ \midrule
         \multirow{2}{*}{MNIST}     & $\ell_2$  & 5.2 & 8.1 & 10.3 & 3.7 & 5.4 & 7.8 & 3.2 & 5.5 & 6.5 \\
                                    & $1$-Wass. & 1.0 & 2.2 & 4.2 & 0.5 & 0.8 & 1.5 & 0.3 & 1.0 & 1.6 \\ \midrule
         \multirow{2}{*}{CIFAR-10}  & $\ell_2$  & 41.0 & 61.6 & 82.6 & 38.4 & 54.8 & 71.4 & 14.8 & 25.4 & 31.3\\
                                    & $1$-Wass. & 1.0 & 2.0 & 3.7 & 0.8 & 1.6 & 2.8 & 0.8 & 1.6 & 2.2 \\ \bottomrule
    \end{tabular}
    \caption{The change under $\ell_p$ or $1$-Wasserstein after applying fixed common perturbations. Results are averaged over the entire test set of either MNIST or CIFAR-10. Gaussian blur is parameterized by the width of the kernel in pixels. The numerical values of the two distance metrics are not directly comparable. Translation is based on the width of the image, which is 28 pixels for MNIST and 32 pixels for CIFAR-10.}
    \label{tab:distance}
\end{table*}

\subsection{Potential smoothness constraints on the threat model}
\label{app:additional_constraints}
We can directly constrain the total variance of the transport plan $\Pi$:

\begin{align*}
   \sum_{\vec i, \vec j \in \Pi} \Pi_{\vec i\to \vec j} \|\vec j-\vec i\|^2 - \left(\sum_{\vec i, \vec j \in \Pi} \Pi_{\vec i\to \vec j} (\vec j-\vec i)\right)^2
\end{align*}

Here, $\vec i$ and $\vec j$ are the position of pixels.
The challenge for this approach is that this total variance constraint is not convex, and therefore cannot be easily integrated into the existing optimization framework.

A more ad-hoc approach is to use only the low-frequency gradient information when taking a PGD step.
This can be done through applying a low-pass filter, e.g., a Gaussian filter, to the gradient matrix.
Low-frequency perturbations have been studied under the $\ell_p$ threat model (CITE); the dynamics between the perturbation frequency under the Wasserstein threat model and how close it is to modeling natural perturbations such as translations, rotations, and blurring is an open problem.

Another potential challenge is the use of local transport plans in \citet{wong_wasserstein_2019} for efficiency.
This approximation can fundamentally limit the magnitude of the perturbations we can model; for example, using a 5-by-5 local transport plan limits the translations we can model to at most 2 pixels.

\end{document}